\documentclass[11pt, conference, compsocconf]{IEEEtran} %
\IEEEoverridecommandlockouts

\usepackage[%
 utf8,
]{inputenc}

\usepackage[T1]{fontenc}

\usepackage{color}
\usepackage{booktabs}
\usepackage{tabularx}
\usepackage{placeins}
\usepackage[para]{threeparttable}
\usepackage{multirow}
\usepackage{multicol}

\usepackage{color}
\definecolor{cobalt}{rgb}{0.0, 0.28, 0.67}
\definecolor{darkblue}{rgb}{0.0, 0.0, 0.55}
\definecolor{lightblue}{rgb}{0.0, 0.4, 0.8}
\definecolor{darkgreen}{rgb}{0.00, 0.44, 0.0}
\definecolor{forestgreen}{rgb}{0.0, 0.27, 0.13}
\definecolor{internationalorange}{rgb}{1.0, 0.31, 0.0}

\newcommand{\del}[1]{}   

\newcolumntype{C}{>{\centering\arraybackslash}X}
\newcolumntype{R}{>{\raggedleft\arraybackslash}X}
\newcolumntype{x}[1]{>{\centering\arraybackslash\hspace{0pt}}p{#1}}

\usepackage[noadjust]{cite}

\ifCLASSINFOpdf
  \usepackage[pdftex]{graphicx}
  \graphicspath{{./fig/}}
  \DeclareGraphicsExtensions{.pdf,.jpeg,.png,.jpg}
\else
  \usepackage[dvips]{graphicx}
\fi

\usepackage{amsmath}

\usepackage{amssymb}
\usepackage{amsfonts}
\usepackage{mathrsfs}
\usepackage{xspace}
\usepackage{bm}
\usepackage{upgreek}

\newcommand{\safemath}[2]{\newcommand{#1}{\ensuremath{#2}\xspace}}

\safemath{\bma}{\mathbf{a}}
\safemath{\bmb}{\mathbf{b}}
\safemath{\bmc}{\mathbf{c}}
\safemath{\bmd}{\mathbf{d}}
\safemath{\bme}{\mathbf{e}}
\safemath{\bmf}{\mathbf{f}}
\safemath{\bmg}{\mathbf{g}}
\safemath{\bmh}{\mathbf{h}}
\safemath{\bmi}{\mathbf{i}}
\safemath{\bmj}{\mathbf{j}}
\safemath{\bmk}{\mathbf{k}}
\safemath{\bml}{\mathbf{l}}
\safemath{\bmm}{\mathbf{m}}
\safemath{\bmn}{\mathbf{n}}
\safemath{\bmo}{\mathbf{o}}
\safemath{\bmp}{\mathbf{p}}
\safemath{\bmq}{\mathbf{q}}
\safemath{\bmr}{\mathbf{r}}
\safemath{\bms}{\mathbf{s}}
\safemath{\bmt}{\mathbf{t}}
\safemath{\bmu}{\mathbf{u}}
\safemath{\bmv}{\mathbf{v}}
\safemath{\bmw}{\mathbf{w}}
\safemath{\bmx}{\mathbf{x}}
\safemath{\bmy}{\mathbf{y}}
\safemath{\bmz}{\mathbf{z}}
\safemath{\bmzero}{\mathbf{0}}
\safemath{\bmone}{\mathbf{1}}

\bmdefine{\biad}{a}
\bmdefine{\bibd}{b}
\bmdefine{\bicd}{c}
\bmdefine{\bidd}{d}
\bmdefine{\bied}{e}
\bmdefine{\bifd}{f}
\bmdefine{\bigd}{g}
\bmdefine{\bihd}{h}
\bmdefine{\biid}{i}
\bmdefine{\bijd}{j}
\bmdefine{\bikd}{k}
\bmdefine{\bild}{l}
\bmdefine{\bimd}{m}
\bmdefine{\bind}{n}
\bmdefine{\biod}{o}
\bmdefine{\bipd}{p}
\bmdefine{\biqd}{q}
\bmdefine{\bird}{r}
\bmdefine{\bisd}{s}
\bmdefine{\bitd}{t}
\bmdefine{\biud}{u}
\bmdefine{\bivd}{v}
\bmdefine{\biwd}{w}
\bmdefine{\bixd}{x}
\bmdefine{\biyd}{y}
\bmdefine{\bizd}{z}

\bmdefine{\bixid}{\xi}
\bmdefine{\bilambdad}{\lambda}
\bmdefine{\bimud}{\mu}
\bmdefine{\bithetad}{\theta}
\bmdefine{\biphid}{\phi}

\safemath{\bmia}{\biad}
\safemath{\bmib}{\bibd}
\safemath{\bmic}{\bicd}
\safemath{\bmid}{\bidd}
\safemath{\bmie}{\bied}
\safemath{\bmif}{\bifd}
\safemath{\bmig}{\bigd}
\safemath{\bmih}{\bihd}
\safemath{\bmii}{\biid}
\safemath{\bmij}{\bijd}
\safemath{\bmik}{\bikd}
\safemath{\bmil}{\bild}
\safemath{\bmim}{\bimd}
\safemath{\bmin}{\bind}
\safemath{\bmio}{\biod}
\safemath{\bmip}{\bipd}
\safemath{\bmiq}{\biqd}
\safemath{\bmir}{\bird}
\safemath{\bmis}{\bisd}
\safemath{\bmit}{\bitd}
\safemath{\bmiu}{\biud}
\safemath{\bmiv}{\bivd}
\safemath{\bmiw}{\biwd}
\safemath{\bmix}{\bixd}
\safemath{\bmiy}{\biyd}
\safemath{\bmiz}{\bizd}

\safemath{\bmxi}{\bixid}
\safemath{\bmlambda}{\bilambdad}
\safemath{\bmmu}{\bimud}
\safemath{\bmtheta}{\bithetad}
\safemath{\bmphi}{\biphid}

\safemath{\bA}{\mathbf{A}}
\safemath{\bB}{\mathbf{B}}
\safemath{\bC}{\mathbf{C}}
\safemath{\bD}{\mathbf{D}}
\safemath{\bE}{\mathbf{E}}
\safemath{\bF}{\mathbf{F}}
\safemath{\bG}{\mathbf{G}}
\safemath{\bH}{\mathbf{H}}
\safemath{\bI}{\mathbf{I}}
\safemath{\bJ}{\mathbf{J}}
\safemath{\bK}{\mathbf{K}}
\safemath{\bL}{\mathbf{L}}
\safemath{\bM}{\mathbf{M}}
\safemath{\bN}{\mathbf{N}}
\safemath{\bO}{\mathbf{O}}
\safemath{\bP}{\mathbf{P}}
\safemath{\bQ}{\mathbf{Q}}
\safemath{\bR}{\mathbf{R}}
\safemath{\bS}{\mathbf{S}}
\safemath{\bT}{\mathbf{T}}
\safemath{\bU}{\mathbf{U}}
\safemath{\bV}{\mathbf{V}}
\safemath{\bW}{\mathbf{W}}
\safemath{\bX}{\mathbf{X}}
\safemath{\bY}{\mathbf{Y}}
\safemath{\bZ}{\mathbf{Z}}

\safemath{\bZero}{\mathbf{0}}
\safemath{\bOne}{\mathbf{1}}
\safemath{\bDelta}{\mathbf{\Delta}}
\safemath{\bLambda}{\mathbf{\UpLambda}}
\safemath{\bPhi}{\mathbf{\Upphi}}
\safemath{\bSigma}{\mathbf{\Upsigma}}
\safemath{\bOmega}{\mathbf{\Upomega}}
\safemath{\bTheta}{\mathbf{\Uptheta}}

\bmdefine{\biAd}{A}
\bmdefine{\biBd}{B}
\bmdefine{\biCd}{C}
\bmdefine{\biDd}{D}
\bmdefine{\biEd}{E}
\bmdefine{\biFd}{F}
\bmdefine{\biGd}{G}
\bmdefine{\biHd}{H}
\bmdefine{\biId}{I}
\bmdefine{\biJd}{J}
\bmdefine{\biKd}{K}
\bmdefine{\biLd}{L}
\bmdefine{\biMd}{M}
\bmdefine{\biOd}{N}
\bmdefine{\biPd}{O}
\bmdefine{\biQd}{P}
\bmdefine{\biRd}{R}
\bmdefine{\biSd}{S}
\bmdefine{\biTd}{T}
\bmdefine{\biUd}{U}
\bmdefine{\biVd}{V}
\bmdefine{\biWd}{W}
\bmdefine{\biXd}{X}
\bmdefine{\biYd}{Y}
\bmdefine{\biZd}{Z}

\bmdefine{\biDelta}{\Delta}
\bmdefine{\biLambda}{\Lambda}
\bmdefine{\biPhi}{\Phi}
\bmdefine{\biSigma}{\Sigma}
\bmdefine{\biOmega}{\Omega}
\bmdefine{\biTheta}{\Theta}

\safemath{\bimA}{\biAd}
\safemath{\bimB}{\biBd}
\safemath{\bimC}{\biCd}
\safemath{\bimD}{\biDd}
\safemath{\bimE}{\biEd}
\safemath{\bimF}{\biFd}
\safemath{\bimG}{\biGd}
\safemath{\bimH}{\biHd}
\safemath{\bimI}{\biId}
\safemath{\bimJ}{\biJd}
\safemath{\bimK}{\biKd}
\safemath{\bimL}{\biLd}
\safemath{\bimM}{\biMd}
\safemath{\bimN}{\biNd}
\safemath{\bimO}{\biOd}
\safemath{\bimP}{\biPd}
\safemath{\bimQ}{\biQd}
\safemath{\bimR}{\biRd}
\safemath{\bimS}{\biSd}
\safemath{\bimT}{\biTd}
\safemath{\bimU}{\biUd}
\safemath{\bimV}{\biVd}
\safemath{\bimW}{\biWd}
\safemath{\bimX}{\biXd}
\safemath{\bimY}{\biYd}
\safemath{\bimZ}{\biZd}

\safemath{\bimDelta}{\biDelta}
\safemath{\bimLambda}{\biLambda}
\safemath{\bimPhi}{\biPhi}
\safemath{\bimSigma}{\biSigma}
\safemath{\bimOmega}{\biOmega}
\safemath{\bimTheta}{\biTheta}

\safemath{\setA}{\mathcal{A}}
\safemath{\setB}{\mathcal{B}}
\safemath{\setC}{\mathcal{C}}
\safemath{\setD}{\mathcal{D}}
\safemath{\setE}{\mathcal{E}}
\safemath{\setF}{\mathcal{F}}
\safemath{\setG}{\mathcal{G}}
\safemath{\setH}{\mathcal{H}}
\safemath{\setI}{\mathcal{I}}
\safemath{\setJ}{\mathcal{J}}
\safemath{\setK}{\mathcal{K}}
\safemath{\setL}{\mathcal{L}}
\safemath{\setM}{\mathcal{M}}
\safemath{\setN}{\mathcal{N}}
\safemath{\setO}{\mathcal{O}}
\safemath{\setP}{\mathcal{P}}
\safemath{\setQ}{\mathcal{Q}}
\safemath{\setR}{\mathcal{R}}
\safemath{\setS}{\mathcal{S}}
\safemath{\setT}{\mathcal{T}}
\safemath{\setU}{\mathcal{U}}
\safemath{\setV}{\mathcal{V}}
\safemath{\setW}{\mathcal{W}}
\safemath{\setX}{\mathcal{X}}
\safemath{\setY}{\mathcal{Y}}
\safemath{\setZ}{\mathcal{Z}}
\safemath{\emptySet}{\varnothing}

\safemath{\colA}{\mathscr{A}}
\safemath{\colB}{\mathscr{B}}
\safemath{\colC}{\mathscr{C}}
\safemath{\colD}{\mathscr{D}}
\safemath{\colE}{\mathscr{E}}
\safemath{\colF}{\mathscr{F}}
\safemath{\colG}{\mathscr{G}}
\safemath{\colH}{\mathscr{H}}
\safemath{\colI}{\mathscr{I}}
\safemath{\colJ}{\mathscr{J}}
\safemath{\colK}{\mathscr{K}}
\safemath{\colL}{\mathscr{L}}
\safemath{\colM}{\mathscr{M}}
\safemath{\colN}{\mathscr{N}}
\safemath{\colO}{\mathscr{O}}
\safemath{\colP}{\mathscr{P}}
\safemath{\colQ}{\mathscr{Q}}
\safemath{\colR}{\mathscr{R}}
\safemath{\colS}{\mathscr{S}}
\safemath{\colT}{\mathscr{T}}
\safemath{\colU}{\mathscr{U}}
\safemath{\colV}{\mathscr{V}}
\safemath{\colW}{\mathscr{W}}
\safemath{\colX}{\mathscr{X}}
\safemath{\colY}{\mathscr{Y}}
\safemath{\colZ}{\mathscr{Z}}

\safemath{\opA}{\mathbb{A}}
\safemath{\opB}{\mathbb{B}}
\safemath{\opC}{\mathbb{C}}
\safemath{\opD}{\mathbb{D}}
\safemath{\opE}{\mathbb{E}}
\safemath{\opF}{\mathbb{F}}
\safemath{\opG}{\mathbb{G}}
\safemath{\opH}{\mathbb{H}}
\safemath{\opI}{\mathbb{I}}
\safemath{\opJ}{\mathbb{J}}
\safemath{\opK}{\mathbb{K}}
\safemath{\opL}{\mathbb{L}}
\safemath{\opM}{\mathbb{M}}
\safemath{\opN}{\mathbb{N}}
\safemath{\opO}{\mathbb{O}}
\safemath{\opP}{\mathbb{P}}
\safemath{\opQ}{\mathbb{Q}}
\safemath{\opR}{\mathbb{R}}
\safemath{\opS}{\mathbb{S}}
\safemath{\opT}{\mathbb{T}}
\safemath{\opU}{\mathbb{U}}
\safemath{\opV}{\mathbb{V}}
\safemath{\opW}{\mathbb{W}}
\safemath{\opX}{\mathbb{X}}
\safemath{\opY}{\mathbb{Y}}
\safemath{\opZ}{\mathbb{Z}}
\safemath{\opZero}{\mathbb{O}}
\safemath{\identityop}{\opI}

\safemath{\veca}{\bma}
\safemath{\vecb}{\bmb}
\safemath{\vecc}{\bmc}
\safemath{\vecd}{\bmd}
\safemath{\vece}{\bme}
\safemath{\vecf}{\bmf}
\safemath{\vecg}{\bmg}
\safemath{\vech}{\bmh}
\safemath{\veci}{\bmi}
\safemath{\vecj}{\bmj}
\safemath{\veck}{\bmk}
\safemath{\vecl}{\bml}
\safemath{\vecm}{\bmm}
\safemath{\vecn}{\bmn}
\safemath{\veco}{\bmo}
\safemath{\vecp}{\bmp}
\safemath{\vecq}{\bmq}
\safemath{\vecr}{\bmr}
\safemath{\vecs}{\bms}
\safemath{\vect}{\bmt}
\safemath{\vecu}{\bmu}
\safemath{\vecv}{\bmv}
\safemath{\vecw}{\bmw}
\safemath{\vecx}{\bmx}
\safemath{\vecy}{\bmy}
\safemath{\vecz}{\bmz}

\safemath{\veczero}{\bmzero}
\safemath{\vecone}{\bmone}
\safemath{\vecxi}{\bmxi}
\safemath{\veclambda}{\bmlambda}
\safemath{\vecmu}{\bmmu}
\safemath{\vectheta}{\bmtheta}
\safemath{\vecphi}{\bmphi}

\safemath{\matA}{\bA}
\safemath{\matB}{\bB}
\safemath{\matC}{\bC}
\safemath{\matD}{\bD}
\safemath{\matE}{\bE}
\safemath{\matF}{\bF}
\safemath{\matG}{\bG}
\safemath{\matH}{\bH}
\safemath{\matI}{\bI}
\safemath{\matJ}{\bJ}
\safemath{\matK}{\bK}
\safemath{\matL}{\bL}
\safemath{\matM}{\bM}
\safemath{\matN}{\bN}
\safemath{\matO}{\bO}
\safemath{\matP}{\bP}
\safemath{\matQ}{\bQ}
\safemath{\matR}{\bR}
\safemath{\matS}{\bS}
\safemath{\matT}{\bT}
\safemath{\matU}{\bU}
\safemath{\matV}{\bV}
\safemath{\matW}{\bW}
\safemath{\matX}{\bX}
\safemath{\matY}{\bY}
\safemath{\matZ}{\bZ}
\safemath{\matzero}{\bmzero}

\safemath{\matDelta}{\bDelta}
\safemath{\matLambda}{\bLambda}
\safemath{\matPhi}{\bPhi}
\safemath{\matSigma}{\bSigma}
\safemath{\matOmega}{\bOmega}
\safemath{\matTheta}{\bTheta}

\safemath{\matidentity}{\matI}
\safemath{\matone}{\bmone}

\safemath{\rnda}{A}
\safemath{\rndb}{B}
\safemath{\rndc}{C}
\safemath{\rndd}{D}
\safemath{\rnde}{E}
\safemath{\rndf}{F}
\safemath{\rndg}{G}
\safemath{\rndh}{H}
\safemath{\rndi}{I}
\safemath{\rndj}{J}
\safemath{\rndk}{K}
\safemath{\rndl}{L}
\safemath{\rndm}{M}
\safemath{\rndn}{N}
\safemath{\rndo}{O}
\safemath{\rndp}{P}
\safemath{\rndq}{Q}
\safemath{\rndr}{R}
\safemath{\rnds}{S}
\safemath{\rndt}{T}
\safemath{\rndu}{U}
\safemath{\rndv}{V}
\safemath{\rndw}{W}
\safemath{\rndx}{X}
\safemath{\rndy}{Y}
\safemath{\rndz}{Z}

\safemath{\rveca}{\bimA}
\safemath{\rvecb}{\bimB}
\safemath{\rvecc}{\bimC}
\safemath{\rvecd}{\bimD}
\safemath{\rvece}{\bimE}
\safemath{\rvecf}{\bimF}
\safemath{\rvecg}{\bimG}
\safemath{\rvech}{\bimH}
\safemath{\rveci}{\bimI}
\safemath{\rvecj}{\bimJ}
\safemath{\rveck}{\bimK}
\safemath{\rvecl}{\bimL}
\safemath{\rvecm}{\bimM}
\safemath{\rvecn}{\bimN}
\safemath{\rveco}{\bomO}
\safemath{\rvecp}{\bimP}
\safemath{\rvecq}{\bimQ}
\safemath{\rvecr}{\bimR}
\safemath{\rvecs}{\bimS}
\safemath{\rvect}{\bimT}
\safemath{\rvecu}{\bimU}
\safemath{\rvecv}{\bimV}
\safemath{\rvecw}{\bimW}
\safemath{\rvecx}{\bimX}
\safemath{\rvecy}{\bimY}
\safemath{\rvecz}{\bimZ}

\safemath{\rvecxi}{\bmxi}
\safemath{\rveclambda}{\bmlambda}
\safemath{\rvecmu}{\bmmu}
\safemath{\rvectheta}{\bmtheta}
\safemath{\rvecphi}{\bmphi}

\safemath{\rmatA}{\bimA}
\safemath{\rmatB}{\bimB}
\safemath{\rmatC}{\bimC}
\safemath{\rmatD}{\bimD}
\safemath{\rmatE}{\bimE}
\safemath{\rmatF}{\bimF}
\safemath{\rmatG}{\bimG}
\safemath{\rmatH}{\bimH}
\safemath{\rmatI}{\bimI}
\safemath{\rmatJ}{\bimJ}
\safemath{\rmatK}{\bimK}
\safemath{\rmatL}{\bimL}
\safemath{\rmatM}{\bimM}
\safemath{\rmatN}{\bimN}
\safemath{\rmatO}{\bimO}
\safemath{\rmatP}{\bimP}
\safemath{\rmatQ}{\bimQ}
\safemath{\rmatR}{\bimR}
\safemath{\rmatS}{\bimS}
\safemath{\rmatT}{\bimT}
\safemath{\rmatU}{\bimU}
\safemath{\rmatV}{\bimV}
\safemath{\rmatW}{\bimW}
\safemath{\rmatX}{\bimX}
\safemath{\rmatY}{\bimY}
\safemath{\rmatZ}{\bimZ}

\safemath{\rmatDelta}{\bimDelta}
\safemath{\rmatLambda}{\bimLambda}
\safemath{\rmatPhi}{\bimPhi}
\safemath{\rmatSigma}{\bimSigma}
\safemath{\rmatOmega}{\bimOmega}
\safemath{\rmatTheta}{\bimTheta}

\usepackage{amssymb}
\usepackage{amsfonts}
\usepackage{mathrsfs}
\usepackage{xspace}
\usepackage{bm}

\newenvironment{textbmatrix}{	\setlength{\arraycolsep}{2.5pt}%
								\big[\begin{matrix}}{\end{matrix}\big]%
								\raisebox{0.08ex}{\vphantom{M}}}

\def\be{\begin{equation}}
\def\ee{\end{equation}}
\def\ben{\nonumber \begin{equation}}
\def\een{\nonumber \end{equation}}
\def\mat{\begin{bmatrix}}
\def\emat{\end{bmatrix}}
\def\btm{\begin{textbmatrix}}
\def\etm{\end{textbmatrix}}

\def\ba#1\ea{\begin{align}#1\end{align}}
\def\bas#1\eas{\begin{align*}#1\end{align*}}
\def\bs#1\es{\begin{split}#1\end{split}} 
\def\bg#1\eg{\begin{gather}#1\end{gather}} 
\def\bi#1\ei{\begin{itemize}#1\end{itemize}}

\safemath{\dirac}{\delta}					
\safemath{\krond}{\dirac}					

\safemath{\upto}{\uparrow}
\safemath{\downto}{\downarrow}
\safemath{\iu}{j}							
\safemath{\ev}{\lambda}						
\safemath{\hilseqspace}{l^{2}}				
\newcommand{\banachfunspace}[1]{\setL^{#1}}	
\safemath{\hilfunspace}{\banachfunspace{2}}	

\safemath{\SNR}{\text{\sc snr}} 				
\safemath{\No}{N_0}							
\safemath{\Es}{E_s}							
\safemath{\Eb}{E_b}							
\safemath{\EbNo}{\frac{\Eb}{\No}}
\safemath{\EsNo}{\frac{\Es}{\No}}

\DeclareMathOperator{\CHop}{\ensuremath{\opH}} 
\safemath{\tvir}{\rndh_{\CHop}}				
\safemath{\tvtf}{\rndl_{\CHop}}				
\safemath{\spf}{\rnds_{\CHop}}				
\safemath{\bff}{H_{\CHop}}					

\safemath{\ircf}{r_{h}}						
\safemath{\tftvcf}{r_{s}}					
\safemath{\tfcf}{r_{l}}						
\safemath{\bfcf}{r_{H}}						

\safemath{\tcorr}{c_h}						
\safemath{\scf}{c_{s}}						
\safemath{\tfcorr}{c_{l}}					
\safemath{\fcorr}{c_{H}}						

\safemath{\mi}{I}							
\safemath{\capacity}{C}						

\safemath{\normal}{\mathcal{N}}			
\safemath{\jpg}{\mathcal{CN}}			
\safemath{\mchain}{\leftrightarrow}		

\safemath{\dB}{\,\mathrm{dB}}
\safemath{\dBm}{\,\mathrm{dBm}}
\safemath{\Hz}{\,\mathrm{Hz}}
\safemath{\kHz}{\,\mathrm{kHz}}
\safemath{\MHz}{\,\mathrm{MHz}}
\safemath{\GHz}{\,\mathrm{GHz}}
\safemath{\s}{\,\mathrm{s}}
\safemath{\ms}{\,\mathrm{ms}}
\safemath{\mus}{\,\mathrm{\mu s}}
\safemath{\ns}{\,\mathrm{ns}}
\safemath{\meter}{\,\mathrm{m}}
\safemath{\mm}{\,\mathrm{mm}}
\safemath{\cm}{\,\mathrm{cm}}
\safemath{\m}{\,\mathrm{m}}
\safemath{\W}{\,\mathrm{W}}
\safemath{\J}{\,\mathrm{J}}
\safemath{\K}{\,\mathrm{K}}
\safemath{\bit}{\,\mathrm{bit}}

\safemath{\define}{=}			

\safemath{\equivalent}{\sim}
\safemath{\distas}{\sim}					
\safemath{\sdiff}{\Delta}				

\safemath{\reals}{\mathbb{R}}
\safemath{\positivereals}{\reals_{+}}
\safemath{\integers}{\mathbb{Z}}
\safemath{\posint}{\integers_{+}}
\safemath{\naturals}{\mathbb{N}}
\safemath{\posnaturals}{\naturals_{+}}
\safemath{\complexset}{\mathbb{C}}
\safemath{\rationals}{\mathbb{Q}}

\usepackage{algpseudocode}

\usepackage{array}

\ifCLASSOPTIONcompsoc
  \usepackage[caption=false,font=normalsize,labelfont=sf,textfont=sf]{subfig}
\else
  \usepackage[caption=false,font=footnotesize]{subfig}
\fi

\usepackage[colorlinks=true,urlcolor=blue]{hyperref}

\ifCLASSINFOpdf
\else
\fi

\begin{document}

\title{Efficient Image Dataset Classification Difficulty Estimation for Predicting Deep-Learning Accuracy
\thanks{IBM, the IBM~logo, and ibm.com are trademarks or registered trademarks of International Business Machines Corporation in the United States, other countries, or both. Other product and service names might be trademarks of IBM or other companies. Submitted. Copyright 2018 by the author(s).}}

\author{
\IEEEauthorblockN{
Florian Scheidegger\IEEEauthorrefmark{1}\IEEEauthorrefmark{2},
Roxana Istrate\IEEEauthorrefmark{3}\IEEEauthorrefmark{2},
Giovanni Mariani\IEEEauthorrefmark{2}
}
\IEEEauthorblockN{
Luca Benini\IEEEauthorrefmark{1}\IEEEauthorrefmark{4},
Costas Bekas\IEEEauthorrefmark{2},
Cristiano Malossi\IEEEauthorrefmark{2}
}
\IEEEauthorblockA{
\IEEEauthorrefmark{1}ETH Zürich, Switzerland
\IEEEauthorrefmark{2}IBM Research - Zürich, Switzerland
}
\IEEEauthorblockA{
\IEEEauthorrefmark{3}Queen's University of Belfast, United Kingdom
\IEEEauthorrefmark{4}Università di Bologna, Italy
}
}

\maketitle

\begin{abstract}
In the deep-learning community new algorithms are published at an incredible pace. Therefore, solving an image classification problem for new datasets becomes a challenging task, as it requires to re-evaluate published algorithms and their different configurations in order to find a close to optimal classifier. To facilitate this process, before biasing our decision towards a class of neural networks or running an expensive search over the network space, we propose to estimate the classification difficulty of the dataset. Our method computes a single number that characterizes the dataset difficulty $27\times$ faster than training state-of-the-art networks. The proposed method can be used in combination with network topology and hyper-parameter search optimizers to efficiently drive the search towards promising neural-network configurations.
\end{abstract}

\section{Introduction}
\label{intro}

Convolutional Neural Networks (CNNs) gained popularity in recent years thanks to the availability of powerful GPUs that enable to efficiently train accurate classification models \cite{he2015delving}. For building practical applications, the deep-learning community shares a common interest in reducing the development cycle, while increasing model accuracy and keeping infrastructure and power consumption expenditure under control. Many publications address these conflicting goals \cite{courbariaux2016binarized}, \cite{gupta2015deep}, \cite{anytimeNeuralNetwork}. Most machine-learning approaches require a human in the loop responsible for taking crucial decisions such as defining the network, finding good combinations of hyper-parameters and performing adequate preprocessing on the input data. To overcome the problem of manual selection various automated approaches such as Grid Search, Random Search \cite{bergstra2012random}, Bayesian optimization \cite{NIPS2012_4522_Bayesian} or Hyperband optimization \cite{li2016hyperband} have been proposed. These methods operate autonomously and improve model performance, however they still have two limiting factors. First, they require a definition of the search space. Second, they consume a large amount of resources for a single optimization task.

In this paper we propose automated methods for quantifying the \textit{difficulty} of a classification problem in terms of how hard it is to reach high accuracy for a given dataset. The proposed method can be used in combination with architecture search optimizers to efficiently drive the search towards promising configurations, avoiding the exploration of unsuitable networks. Consciously or not, the characterization of dataset difficulty is a process followed by every deep learning architect. When looking for a well-performing model for a new dataset, common practice is to try state-of-the-art networks to evaluate how hard is to classify the images in the dataset. Since datasets are large and models complex, the process of training, comparing, and selecting a few state-of-the-art deep networks becomes a computationally heavy task. We propose to optimize this step by providing a classification difficulty estimator, that provides insights into the classification task and can be used to rapidly confine the exploration to a few promising networks. We aim to construct dataset characterizations that run orders of magnitude faster than the actual training and have high correlation with state-of-the-art network accuracies.

In summary, our main contributions are the following:

\begin{itemize}
    \item We propose and evaluate three different dataset complexity scoring pipelines.
    \item We conduct various deep learning experiments with fixed hyper-parameter and data augmentation configurations that run on thirteen datasets. 
    \item We evaluate approximate computing techniques, such as subsampling and early stopping, in order to reduce the execution time without affecting the end results.
\end{itemize}

The remainder of the paper is organized as follows. Section~\ref{sec:rw} describes the related work, Section~\ref{sec:not} introduces the notation used throughout the paper, Section~\ref{sec:ranking_ds} details the adopted methodologies, Section~\ref{sec:results} examines the results, and Section~\ref{sec:conclusion} concludes the current work.

\section{Related Work}
\label{sec:rw}
The topic of difficulty estimation of a dataset is scarcely explored in the literature. In \cite{tudor2016hard}, the authors address this problem by using, but use as reference the human response time for solving a visual search task.  As compared to our technique which focuses on defining how easily separable are the different classes in a dataset, their technique analyses a dataset difficulty on an image based approach and employs two VGG-like \cite{vgg} networks that work as encoders and extract features that are further passed through a regressor. The complexity of this solution comes from passing the full dataset through the VGG-like networks, since the VGG family includes one of the largest state-of-the-art network with 138M parameters and 15 GFLOPs/inference.

\begin{figure}[t]
\centering
\includegraphics[width=1\linewidth]{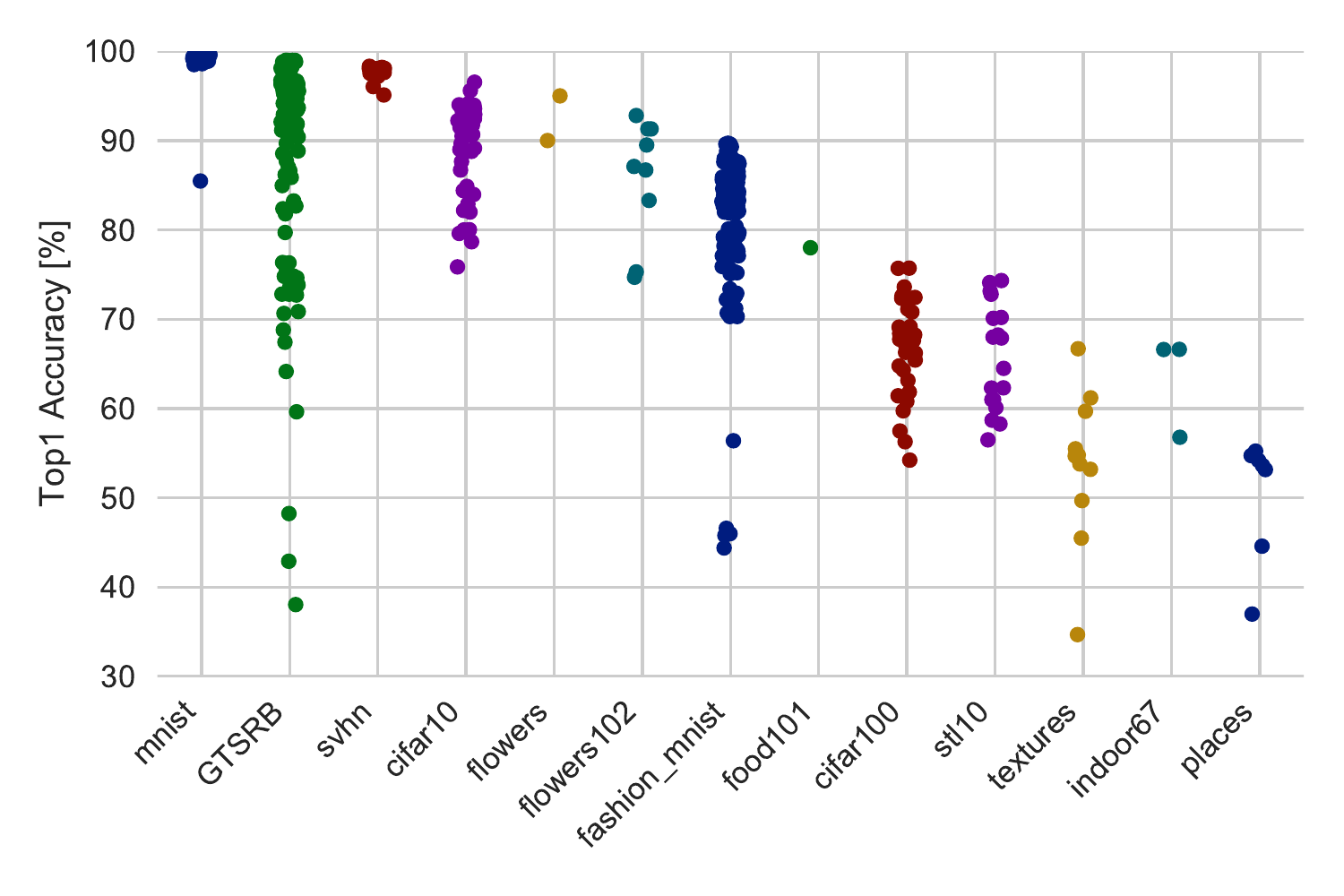} 

\caption{State-of-the-art results achieved on various datasets. Each point corresponds to a single published algorithm. For well established datasets, or for datasets that were part of a public challenge, many Top1 results are known. The picture combines vanilla CNNs with or without transfer learning, non-deep learning approaches such as SVMs or Random Forests applied on handcrafted features, and other problem specific pipelines.}
\label{fig:GT_sets001}
\end{figure}

Figure~\ref{fig:GT_sets001} shows state-of-the-art accuracies achieved by thirteen of the most commonly used image classification datasets: \textit{MNIST} \cite{data_mnist}, \textit{GTSRB} \cite{data_gtsrb}, \textit{svhn} \cite{data_svhn}, \textit{CIFAR10} \cite{data_cifar10_100}, \textit{flowers}\footnotemark\footnotetext{Available at \url{http://download.tensorflow.org/example_images/flower_photos.tgz}}, \textit{flowers102} \cite{nilsback2008_flowers102}, \textit{fashion MNIST} \cite{data_fashion_mnist}, \textit{food101} \cite{bossard2014_food}, \textit{CIFAR100} \cite{data_cifar10_100}, \textit{stl10} \cite{data_stl10}, \textit{textures} \cite{cimpoi2014_textures}, \textit{indoor67} \cite{quattoni2009_mit67}, and \textit{places} \cite{data_zhou2017places}. As expected, more results are available for the highly curated datasets, such as \textit{MNIST} and \textit{CIFAR10/100}. The same holds for datasets introduced as part of a competition, such as the German Traffic Sign Recognition Benchmark (\textit{GTSRB}), where the authors published the full performance list of over a hundred machine learning algorithms. For less popular datasets, results are harder to acquire. In this category fits the flowers dataset, which was introduced in a tensorflow tutorial\footnotemark\footnotetext{Available at \url{https://www.tensorflow.org/tutorials/image_retraining}} with the purpose of explaining transfer learning. In that case, transferred learning \cite{goodfellow2016deep} provides superior classification performance over any CNN trained from scratch. General embeddings obtained from pretrained models help increasing performance for a specific task, especially when the data is limited \cite{darioPaper}. Figure~\ref{fig:GT_sets001} reveals that all mentioned algorithms, even non-CNN based ones, easily reach accuracies above $95\%$ on \textit{MNIST}. This observation motivated the authors of \textit{fashion MNIST} to introduce a more diverse dataset in terms of images, but equal to \textit{MNIST} in terms of number of training/validation samples, image sizes and number of classes. This allows any algorithm designed for \textit{MNIST} to run without modification on \textit{fashion MNIST}. Their initial study demonstrates a wider spread in performance among different algorithms and consistently lower performance when compared with the same algorithms evaluated on \textit{MNIST} \cite{data_fashion_mnist}.

To the best of our knowledge, there is no published work that focuses on automatically ranking classification difficulty among datasets.

\section{Notation}
\label{sec:not}
In this work, we refer to a dataset with the quadruple $\setD \quad := \quad (\matX_{train}, \vecy_{train}, \matX_{test}, \vecy_{test})$, where $\matX_{train} \in \reals^{n_{train} \times d}$ and $\matX_{test} \in \reals^{n_{test} \times d}$ are the training and testing inputs, and $\vecy_{train} \in [1,C]^{n_{train}}$ and $\vecy_{test} \in [1,C]^{n_{test}}$ are the training and testing labels. We assume that the datasets come already split into train and test sets, as this is commonly the case for published data. We denote the input dimension as $d$, the number of input samples as $n_{train}$ for training and $n_{test}$ for testing, and the number of classes as $C$. $\setM$ refers to a model, including the network topology and related hyper-parameters, and it includes the training and data augmentation related hyper-parameters. Therefore, the tuple $(\setD, \setM)$ specifies a deep learning training run of model $\setM$ on dataset $\setD$. We denote with $\text{Top-1}(\setD, \setM)$ the Top-1 accuracy classification performance of the training run. In all experiments, training is performed with $(\matX_{train}, \vecy_{train})$ and performance is measured on $(\matX_{test}, \vecy_{test})$.

\section{Ranking Datasets}
\begin{figure*}[t]
\centering
\includegraphics[width=1.0\linewidth]{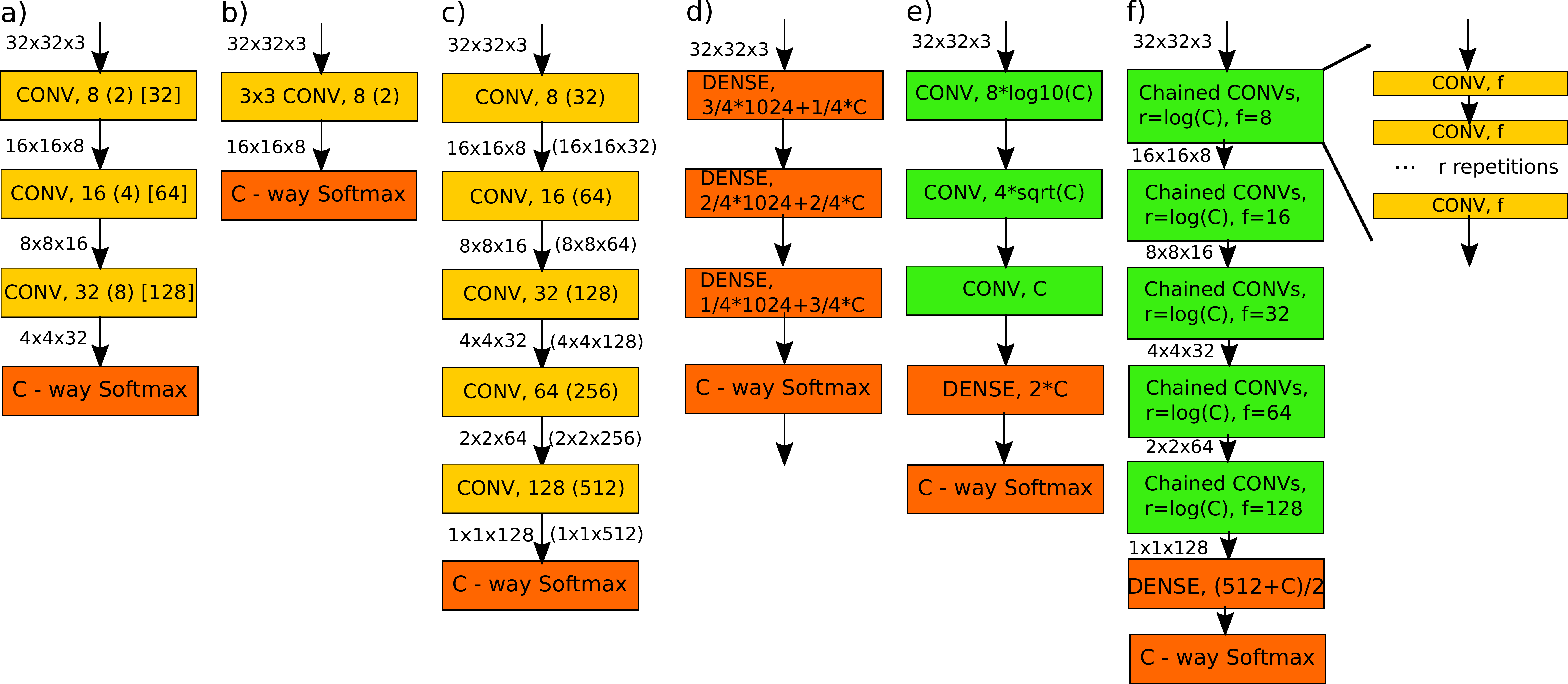} 
\caption{\textit{Probe nets}: simple deep learning architectures used to characterize datasets. \textit{Static} networks are shown in Subfigure a), b) and c), they only differ in the dense layer connections in the softmax output that ends with a dataset specific number of classes $C$. Subfigure a) shows a \textit{regular}, \textit{narrow} and \textit{wide} probe net that differ in kernel depth, b) shows a \textit{shallow} version and c) a \textit{deep} version of the net for non and \textit{normalized} kernel depths. \textit{Dynamic} networks, d), e) and f), scale the topology with respect to the number of classes $C$. Subfigure d) consists of dense layers that scale the number of hidden units according to linear weighted sum between input and output dimension, Subfigure e) scales kernel depths according to $C$ and Subfigure f) scales the number or repetions of stacked static layers according $C$.}
\label{fig:probenet_arch}
\end{figure*}

\label{sec:ranking_ds}
In this work we proposed metrics to quantify the \textit{difficulty} of image classification datasets. We propose dataset scoring functions $r(\setD)$ to map a dataset $\setD$ to a scalar real number with the goal of ranking different datasets in terms of classification accuracy estimates.

\subsection{Silhouette Score}
\label{sec:ss}

The silhouette score is a well established metric that compares tightness of same-class samples to separation of different-class samples \cite{rousseeuw1987silhouettes}. Let $i$ be one input sample, $a(i)$ the average euclidean distance between the sample and all the points $j$ belonging to the same class as $i$, and $b(i)$ the average distance between $i$ and all points $j$ of the closest different class. The \textit{silhouette} of the $i$-th sample is computed \cite{rousseeuw1987silhouettes} as follows:

\begin{equation}
\label{eq:sil}
s(i) \quad := \quad  
\begin{cases} 
1 - a(i)/b(i),  &\text{if}\ a(i) < b(i) \\
0,              &\text{if}\ a(i) = b(i) \\
b(i)/a(i) - 1,  &\text{if}\ a(i) > b(i).
\end{cases}
\end{equation}

The silhouette of one class is defined as the average over all samples belonging to that class and the overall silhouette score of the full dataset is defined as average over all samples. The definition of 
the quantities $a(i)$ and $b(i)$ are based on pairwise distances between two samples $i$ and $j$. The silhouette score complexity is $O(\bar{d}n^2)$, where $n$ is the number of samples and $\bar{d}$
is the cost of computing the distance of one pair of samples as mean squared error (MSE) distance in the $\reals^{\bar{d}}$. Since, the MSE distance in the original domain is a poor measurement for image similarities, we apply first a transformation $\reals^d \rightarrow \reals^{\bar{d}}$ that maps images into a space that better reflects distances between image pairs.

\begin{table}[t]
\small
\begin{threeparttable}
\renewcommand*{\arraystretch}{0.9}
\caption{Pipelines used to compute the silhouette score on datasets}
\label{tab:algo_ss}
\begin{tabularx}{\columnwidth}{@{} c|l|c|l|l @{}}
\toprule
Score & Transformation & $\bar{d}$ & Distance & Speedup  \\
\midrule
$S_1$   & None          & $d$ &     MSE &   $31.3\times $ \\
$S_2$   & None          & $d$ &     DSSIM & $1.0$ (Ref) \\
$S_3$   & Resize image  & $8^2$ &   MSE &   $48.4\times $ \\
$S_4$   & Resize image  & $8^2$ &   DSSIM & $1.3\times $ \\
$S_5$   & PCA           & $10$ &    MSE &   $72.8\times $ \\
$S_6$   & Autoencoder   & $1000$ &   MSE &  $6.4\times $ \\
\bottomrule
\end{tabularx}
\end{threeparttable}
\end{table}

Table~\ref{tab:algo_ss} provides details on the applied pipelines. We decided to include a resizeing of the images to a small resolution of $8\times8$ pixels, applying principal component analysis (PCA) to reduce the dimension to 10, and using a fixed encoding based on a pretrained CNN inference. We considered as encoder a ResNet-50 \cite{he2016deep_resnet} network pretrained on ImageNet \cite{deng2009imagenet} to produce generalized per image feature vectors of dimensionality 1000 by taking the output of the last fully connected layer before applying the non-linearity. Additional to the MSE distance, we used the structural dissimilarity index DSSIM \cite{ssim} to compare images with a metric that captures spatial information. Due the squared complexity, we applied heavy subsampling and run all computations with a maximum of 1000 randomly selected samples, resulting in a distance matrix with at most $1M$ entries. Table~\ref{tab:algo_ss} states timing among the different pipelines. For fast execution, it is crutial to operate in a low dimensional space and to use a simple distance metric.

\subsection{$K$-means Clustering}
\label{sec:algo_cluster}
The complexity of the silhouette scores detailed in Subsection~\ref{sec:ss} scales with $n^2$, and computing it is a slow process even after subsampling. In general, the complexity of a deep-learning job is $O(c(\setM)n_{train}e))$, where $e$ is the number of epochs. During one epoch the full training set consisting of $n_{train}$ samples is fed once with a computational cost of $c(\setM)$, where $c(\setM)$ is a model dependent constant. Even though complex models might have large computational cost of $c(\setM)$, the asymptotic behaviour of a training job is linear in $n$. For this reason, the asymptotic behaviour of the silhouette score computation of $n^2$ is outperformed by the actual training job. Competitive scoring metrics should execute faster than a train job itself, thus we are looking for scores with at most linear complexity in $n$.

We propose to run a (fast) clustering algorithm to produce class labels $\tilde{\vecy}$ and evaluate the full dataset based on metrics that compare $\tilde{\vecy}$ against the ground truth labels $\vecy$. We assess the following known scores: adjusted mutual information \cite{vinh2010information_AMI}, adjusted rand index \cite{hubert1985comparing_adjusted_RI}, completeness, homogeneity and the v-measure \cite{rosenberg2007v}. Additionally, we propose an own tailored score based on the estimation of the confusion matrix built between the cluster indices and the true labels. Clustering algorithms work in an unsupervised fashion, meaning that the clusters are not assigned to any label. To estimate the confusion matrix we require an one-to-one mapping between the clusters and the labels. A naïve solution computes all possible permutations, and selects the one that maximizes the trace normalized by the amount of total data points. Due to the factorial complexity, we propose to compute an approximate estimate with the following greedy algorithm. First, we search the maximum values per row and assign the clusters to the corresponding label. If the greedy strategy resolves to a bijective mapping, the accuracy is computed based on the found confusion matrix. Otherwise, the non-bijective contradictions are solved with extensive search. Since this approach still results in a worst case factorial complexity, we set an maximum limit of contradictions to be solved by exhaustive search to seven, and solve other contradictions by assigning an initial permutation. Hence, the accuracy estimate is the maximum achieved over all values obtained with permutations on the remaining problematic locations. It turns out that the proposed procedure is stable and fast to evaluate and it builds a lower bound to the optimal value. In few cases our chosen maximum limit of seven contradictions was exceeded.

\subsection{Probe Nets}
\label{sec:algo_probe_nets}
As alternative scoring metric we also investigate the possibility of training a small predefined neural network and score the dataset based on its accuracy. We call this network a \textit{probe network}. The probe net model $\setM_{probe}$ must be general enough to be applied to any image classification task and considerably faster: $c(\setM_{probe}) << c(\setM)$. Additionally, to further speedup the execution time we stop the training of the probe net after few epochs, before full convergence.
\begin{table}[t]
\small
\begin{threeparttable}
\renewcommand*{\arraystretch}{0.9}
\caption{Operation count and number of parameters of proposed probe nets}
\label{tab:ProbeNets_exe_performance}
\begin{tabularx}{\columnwidth}{@{} l|rr|rr @{}}
\toprule
Probe Net & \multicolumn{2}{c}{$C=10$} & \multicolumn{2}{c}{$C=100$} \\
\cmidrule(r){2-3}
\cmidrule(r){4-5}
& OPs & Weights & OPs & Weights \\
\midrule
Regular             	&  0.81M	&    11K	&  0.86M	&   57.5K \\
Narrow					&  0.09M	&     2K	&  0.10M	&    13K \\
Wide					& 10.34M	&   114K	& 10.52M	&   299K \\
Shallow					&  0.24M	&    21K	&  0.42M	&   205K \\
Shallow norm.	    	&  0.06M	&     5K	&  0.10M	&    51K \\
Deep					&  1.40M	&   100K	&  1.41M	&   112K \\
Deep norm.	    		& 19.76M	&  1576K	& 19.81M	&  1622K \\
\midrule
MLPs				    & 2.90M	    &  2908K	&  3.10M	&  3107K \\
Kernel depth		    & 0.53M	    &     6K	&  4.56M	&   384K \\
Length  			    & 1.41M	    &   118K	&  4.39M	&   338K \\
\midrule
ResNet-20               & 40.55M	&  271K     & 40.56M	&   277K \\    
\bottomrule
\end{tabularx}
\end{threeparttable}
\end{table}

\begin{figure*}[t]
\centering
\includegraphics[width=1.0\linewidth]{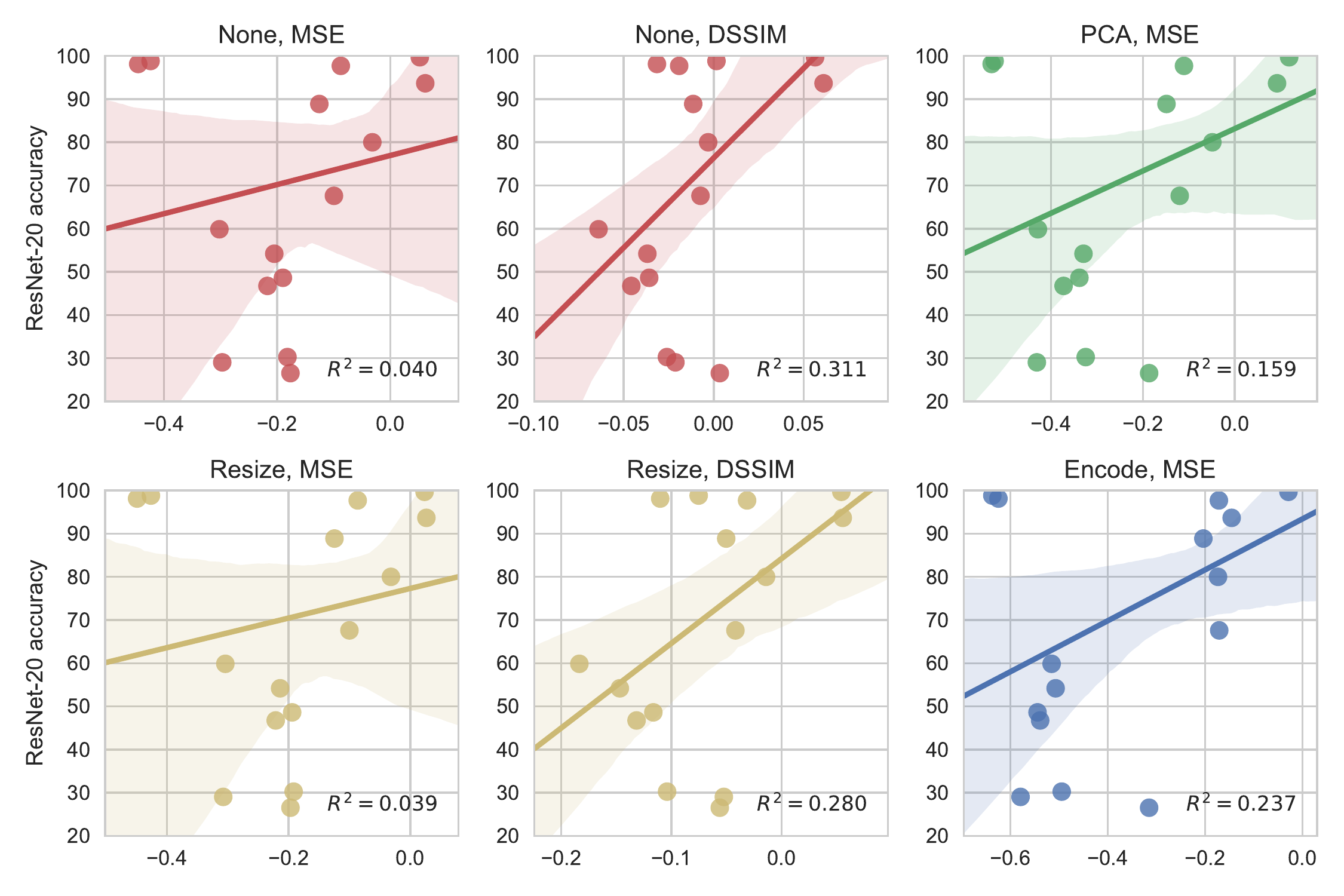} 
\caption{Prediction performance of silhouette scores: the $x$-axis states variations of the silhouette scores according Table~\ref{tab:algo_ss} and the $y$-axis states the model performance obtained by a state-of-the-art architecture, a ResNet-20 topology. Subfigures vary according pretransformation \{None, Resize, PCA, Encode\} and pairwise distance metrics \{MSE, DSSIM\}.}
\label{fig:ss_result}
\end{figure*}

We propose to construct variations of two types of probe nets: \textit{static probe nets} that have a fixed topology and \textit{dynamic probe nets} that scale the topology according to the number of classes. The \textit{regular probe net} consists of three convolutional layers, each followed by batch normalization, max pooling of size $2\times2$, and ReLU activations, which are defined element-wise as $x \mapsto \max(0, x)$. We used eight kernels in the first layer and doubled the number of kernels per layer. We provide \textit{wide} and \textit{narrow} variations that scale the number of kernels per layer up and down by $4\times$, respectively. \textit{Shallow} and \textit{deep} variations are obtained by subtracting and adding two layers, respectively. Since doubling the kernel sizes per layer leads to different tensor shapes between the last convolution and the $C$-way softmax, the \textit{non-normalized} \textit{shallow} and \textit{deep probe nets} have a considerable different number of trainable parameters. We define \textit{normalized probe networks} to match the number of trainable parameters of the output layer of the \textit{regular probe net}. We construct \textit{dynamic nets} with a more complex topology to account for more classes. This is achieved either by scaling dependent on $C$ the number of hidden units in an multilayer percpetron \textit{mlp}, the number of filters (\textit{filter depth scaled probe nets}), or the number of stacked filters (\textit{length scaled probe net}). Figure~\ref{fig:probenet_arch} shows the ten proposed prob net architectures.

\section{Results}
\label{sec:results}
\begin{figure*}[!ht]
\centering
\includegraphics[width=1.0\linewidth]{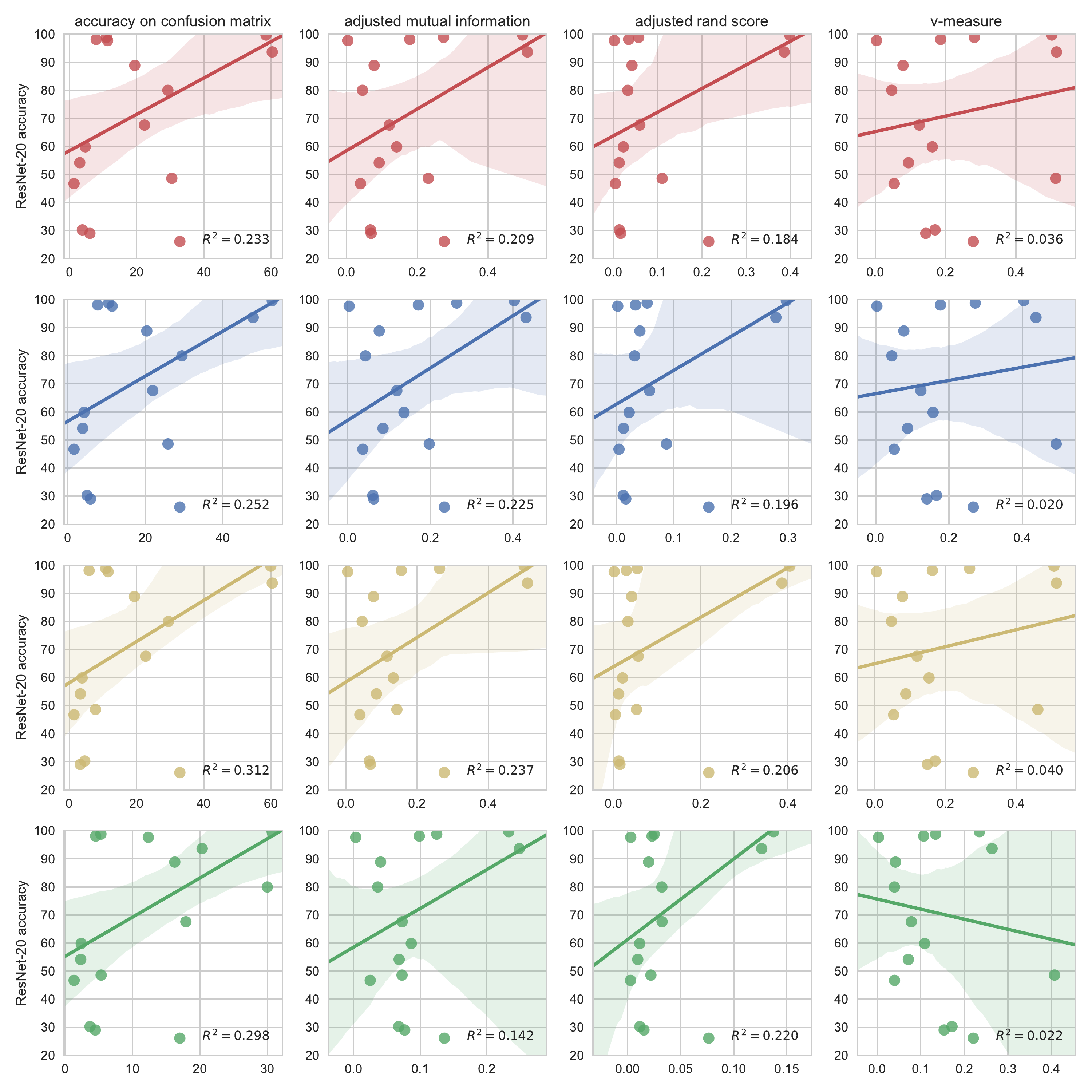} 
\caption{Prediction performance of $k$-means based dataset scores: the $x$-axis states variations of the $k$-means based scores from Subsection~\ref{sec:algo_cluster} and the $y$-axis states the model performance obtained by a state-of-the-art architecture, a ResNet-20 topology. Rows state results for applying none, resizing, PCA, and encoding as pretransformations. Columns state results for four metrics comparing the clustering with the ground truth classification. Scores based on the estimated confusion matrix outperform the others and applying PCA or an embedding is beneficial for accuracy prediction.}
\label{fig:kmeans_result}
\end{figure*}

\begin{figure*}[!t]
\centering
\includegraphics[width=1.0\linewidth]{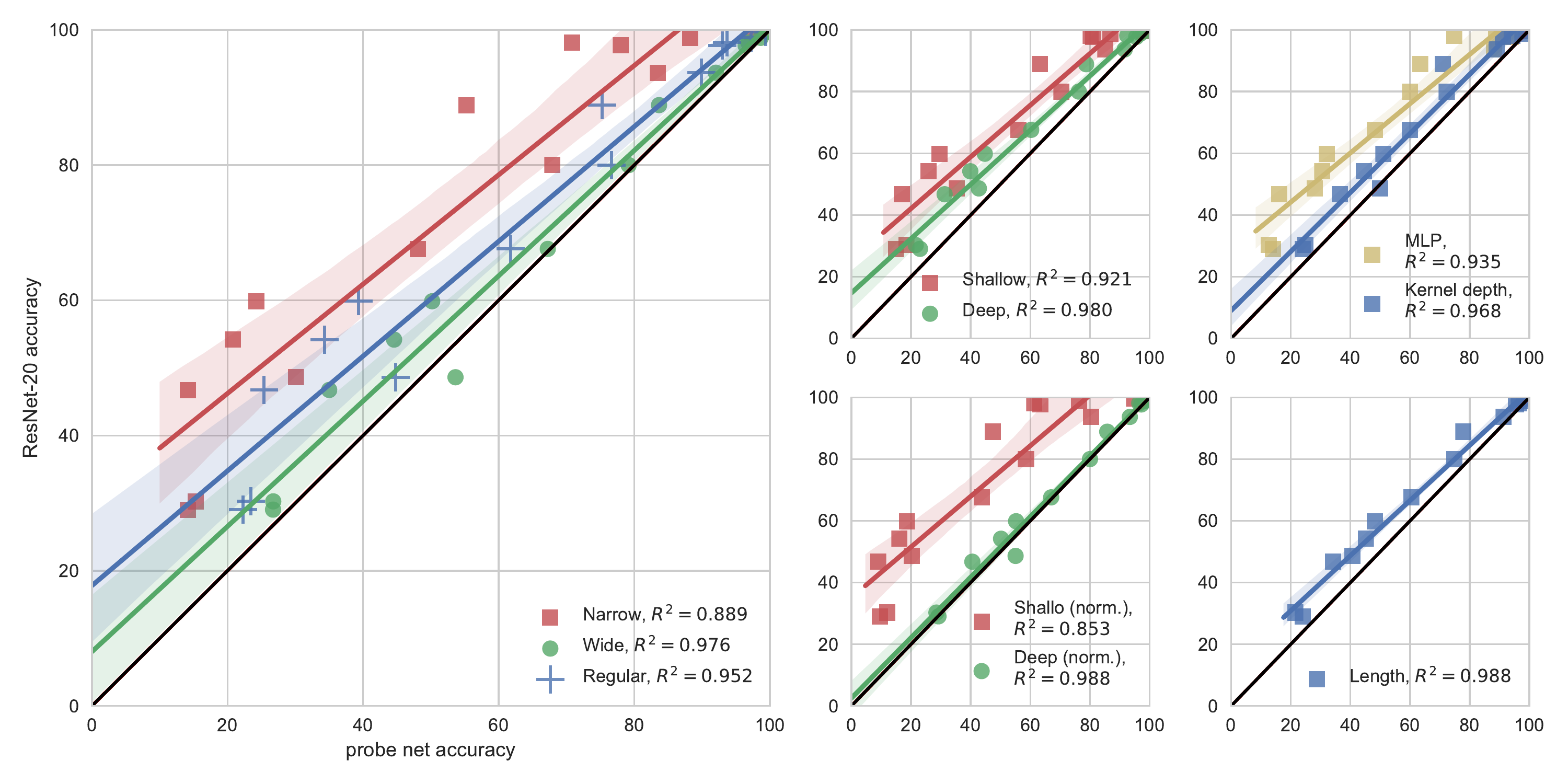} 

\caption{Prediction performance of probe nets: the $x$-axis states the accuracy reached of a converged probe net and the $y$-axis states the model performance obtained by a state-of-the-art architecture, a Resnet-20 topology. All seven static probe nets, a) \textit{regular/narrow/wide}, b) \textit{shallow/deep}, c) \textit{shallow/deep normalized}, and the three dynamic probe nets, d) \textit{mlp}, \textit{kernel depth} scaled, and e) \textit{length} scaled are strongly correlated with results obtained with ResNet-20.}
\label{fig:probenet_result}
\end{figure*}

In order to perform a fair evaluation, we fix hyper-parameters throughout the experiments. We use a ResNet-20 topology to compute the reference accuracy on all datasets resized to $32\times32$ pixels. We follow the data augmentation described in \cite{lee2015deeply}. We use the RMSProp \cite{tieleman2012lecture_rmsprop} optimizer to minimize the average cross entropy with a learning rate of $10^{-4}$. All evaluations employ the He initialization \cite{he2015delving_HeInit} with a gain factor $1.0$ and a constant batch size of $32$. Training is run for $100$ epochs. Results presented in Figure~\ref{fig:ss_result}, Figure~\ref{fig:kmeans_result}, and Figure~\ref{fig:probenet_result} state the computed classification difficulty score on the $x$-axis plotted against the ResNet-20 accuracy reference that is shared among all plots. An ideal dataset difficulty score should obey a linear dependency and match the reference accuracies.

\subsection{Silhouette Score}
\label{subsec:ss_res}
Figure~\ref{fig:ss_result} compares the scores based on the proposed pipelines in Table~\ref{tab:algo_ss} with the reference accuracies of a ResNet-20 \cite{he2016deep_resnet}. The silhouette score based on the DSSIM pairwise distance outperforms the MSE based distance, as the former preserves spatial information of the image domain. Similarly, the (more expensive) computation with the full image domain slightly outperforms the resized counterpart, since it benefits from more information. Subfigure c) shows results when the silhouette score is applied on the PCA reduced space and in Subfigure f) the PCA is replaced by the autoencoder. Although reducing the dimensionality is often beneficial, we obtained the best correlation results with the DSSIM based distance on the original domain, with a weak correlation of $R^2=0.31$.

\begin{figure}[h]
\centering
\includegraphics[width=1.0\linewidth]{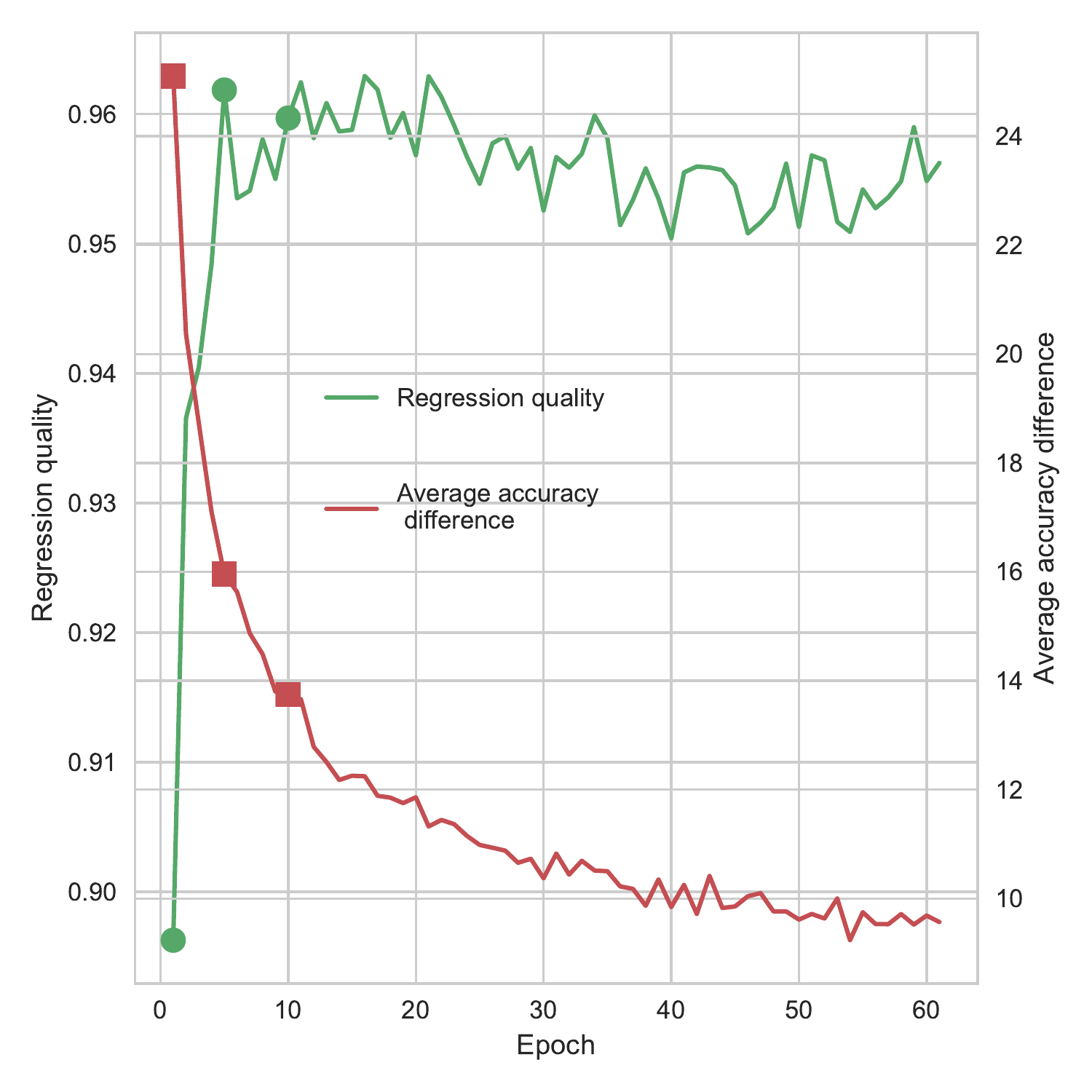} 
\caption{Evolution of the prediction quality over training epochs of the \textit{regular} probe net. The regression quality saturates around $R^2 = 0.96$ within a few epochs while the average accuracy difference between the probe net and the reference is further decreased for longer training.}
\label{fig:probenet_epoch_q}
\end{figure}

\begin{figure}[h]
\centering
\includegraphics[width=1.0\linewidth]{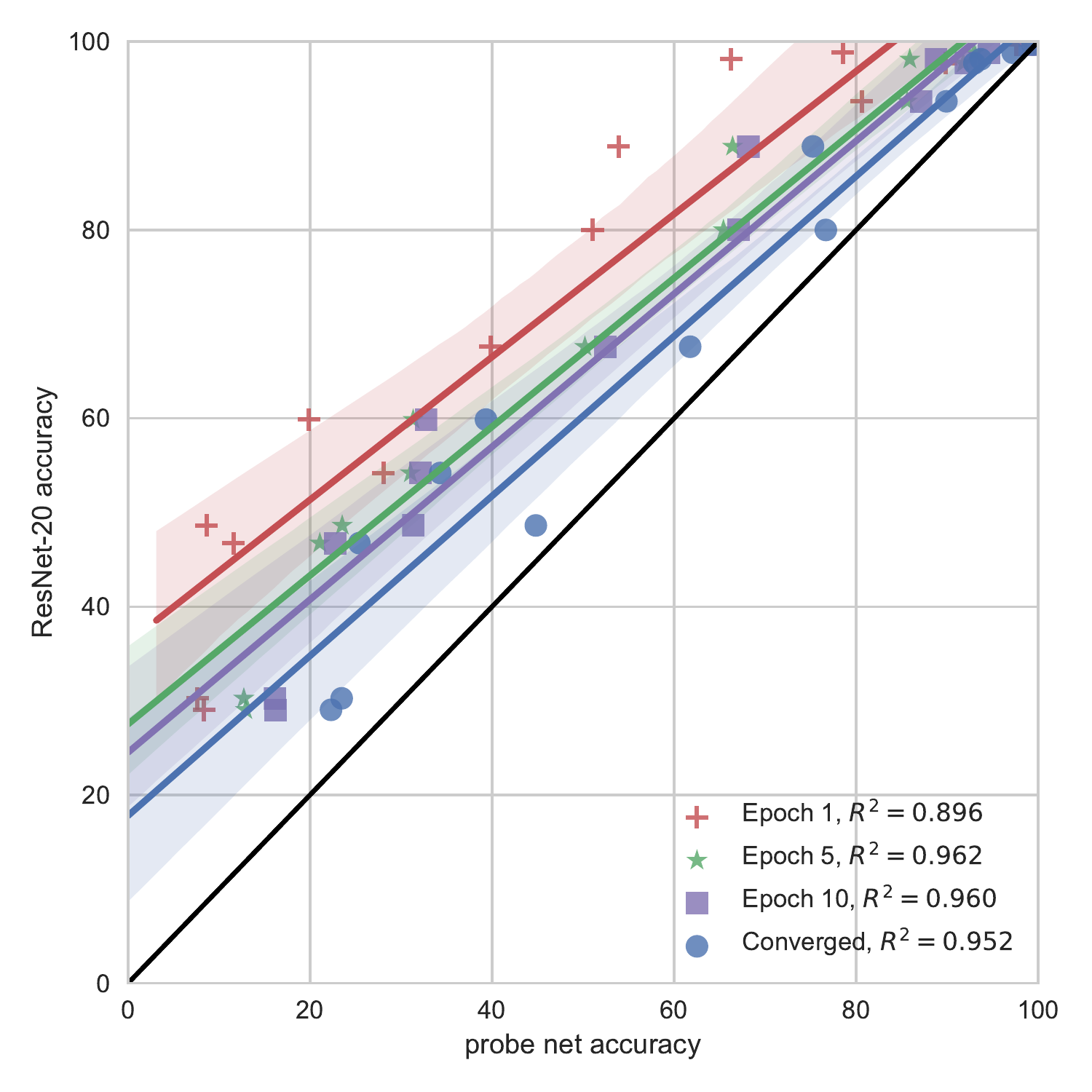} 
\caption{Regression snap shots obtained at epoch 1, 5, 10 and at convergence of the \textit{regular} probe nets.}
\label{fig:probenet_epoch_snapshot}
\end{figure}

\subsection{$K$-means Clustering}
\label{subsec:res_kmeans}
For the evaluation of the proposed $k$-means based scoring pipeline (see Subsection~\ref{sec:algo_cluster}), we cluster the images in $C$ clusters, where $C$ is the knwon number of categories in the dataset. For a faster convergence, we initialize the centroids with the average image of each class. $k$-means runs based on the euclidean $L_2$ distance with a stopping tolerance of $10^{-4}$ and a maximum of $300$ iterations without random restarts. Figure~\ref{fig:kmeans_result} shows the regression among proposed scores and the obtained reference accuracy of a ResNet-20.

Among columns different scores are assessed, such as the \textit{accuracy on the estimated confusion matrix (AECM)}, the \textit{adjusted mutual information} score, the \textit{adjusted random score} and the \textit{v-measure}. Results for the \textit{homogeneity score} and the \textit{completeness score} are omitted since they are highly correlated with the \textit{v-measure}. Except the \textit{v-measure} computed in the encoded setting, the scores are weakly positively correlated with the obtained ResNet-20 reference. Resizing to a small dimension of $8\times8$ only marginally affects results, applying PCA helps to improve predictions from around $R^2=0.25$ up to $R^2=0.31$. The encoder based pipeline provides an embedding that is in the same order of quality ($R^2 = 0.30$). Among the score computations the one based on AECM outperforms the other scores. The weak performance of the $k$-means clustering is due to known limitations, such as no global minimum guarantee and poor distance metric that ignores the spatial information. $k$-means clustering based pipelines are $5.2\times$ (no pretransformation) up to $50.5\times$ (PCA pretransformation) faster than silhouette score based pipelines (comparison includes the faster MSE timings) when comparing execution times in terms of average per input sample.

\subsection{Probe Nets}
\label{subsec:res_probenets}
All proposed probe nets, as presented in Figure~\ref{fig:probenet_arch}, are trained with the same constant configuration and data augmentation parameters as explained in Section~\ref{sec:results}. Results are obtained after training for 100 epochs. Figure~\ref{fig:probenet_result} shows all obtained correlations between runs of the ten proposed probe nets against the reference. All probe nets share a high correlation with the reference ResNet-20 of $R^2 > 0.88$ and consistently outperform results achieved with the $k$-means based approach, see Subsection~\ref{subsec:res_kmeans}. Subfigure a) shows an increasing correlation of $R^2 = 0.89$ to $R^2 = 0.98$ between \textit{narrow}, \textit{regular}, and \textit{wide} probe nets and the reference. This can be explained by the better generalization ability of the network with more degrees of freedom, at the cost of an increased execution time. For more details see Table~\ref{tab:ProbeNets_exe_performance}. \textit{Deep} probe nets topologies outperform their \textit{shallow} counterparts.
This effect is even more prominent in the \textit{normalized} case, Subfigure~b) versus Subfigure~d). We observe that a better generalization performance is mainly driven by a larger amount of tuneable parameters that comes at the cost of increased execution timings. Subfigures~c) and e) of Figure~\ref{fig:probenet_result} show the results for probe nets that dynamically adapt the architecture topology to the number of classes. The dependency of the architecture on the number of classes implies different execution times on datasets with different number of classes. The \textit{mlp} can not compete with the CNN counterparts.  

\subsection{Efficient Evaluation of Probe Nets}
As presented in Subsection~\ref{subsec:res_probenets} probe nets have a good predictive behaviour of what a reference network achieves on a given dataset. However, that information is only valuable if it can be computed order of magnitudes faster than training large models. The way probe nets are constructed give them an inherent computational benefit over the full model. In addition, we exploit early stopping of the learning to further reduce the computational time of the probe net. Note that we can stop the probe net before convergence, since we are interested in the learning trend that characterizes the problem's difficulty, not in the final accuracy. Figure~\ref{fig:probenet_epoch_q} shows how the prediction quality improves for a regular probe net with an increasing amount of epochs for which it is trained on all datasets. Within a few epochs the regression quality reaches a saturation at about $R^2=0.95$. The mean accuracy difference between the probe nets and the reference ResNets (trained till convergence) is further decreased, meaning that the probe nets are not yet converged and are still increasing their own classification performance. Figure~\ref{fig:probenet_epoch_q} highlights the achieved quality at epoch 1, 5 and 10. Figure~\ref{fig:probenet_epoch_snapshot} presents both intermediate and after convergence results. With increasing number of epochs, the regression moves from the top-left corner towards the identity line in the middle. As few as 5 epochs are enough to reach the full dataset performance prediction ability, well before that actual probe net has converged.

\section{Conclusion}
\label{sec:conclusion}
We formulated the question to compute a ranking among datasets that reflect their
inherent classification difficulty. We suggested three processing pipelines, a silhouette based score, a $k$-means clustering based and a probe net based evaluation pipeline. The main drawback of the silhouette based approach is the high complexity, which scales with the squared number of samples. We proposed efficient score computing pipelines based on $k$-means and probe nets that scale linear in the number of samples. $k$-means delivers results one complexity class faster and with similar prediction quality as the silhouette approach, reaching a weak correlation with reference models of $R^2=0.31$. Finally, we presented the probe nets, which are small networks, and apply standard deep learning techniques in order to compute predictions that are strongly correlated with the reference from $R^2=0.89$ up to $R^2=0.99$. Even the worst performing probe net outperforms silhouette and $k$-means based scoring with a wide quality margin. We further evaluated the fact of early stopping to reduce the data score evaluation time and observed little to no performance drop. Leveraging the small architectures of probe nets and early stopping allows to perform dataset scoring $27\times$ faster than the required training time of the actual reference model. 

\section*{Acknowledgements}
We would like to thank Dr. Dario Garcia Gasulla from the Barcelona Supercomputing Center for discussion and advise.
This project has received funding from the European Union’s Horizon 2020 research and innovation programme under grant agreement No 732631, project OPRECOMP.

\newcommand{\BIBdecl}{\setlength{\itemsep}{0 pt}}
\bibliographystyle{IEEEtran}
\bibliography{IEEEabrv,bib}
\end{document}